\documentclass[conference]{IEEEtran}
\makeatletter\if@twocolumn\PassOptionsToPackage{switch}{lineno}\else\fi\makeatother
\usepackage{bigdelim}

\usepackage{graphicx}
\usepackage[T1]{fontenc}
\usepackage[numbers,sort&compress]{natbib}
\usepackage{mathtools}
\usepackage{booktabs}
\usepackage{lineno,hyperref}
\usepackage{amsmath}
\usepackage{amsfonts}
\usepackage{amssymb}
\usepackage{amsmath}
\usepackage{multirow,morefloats,floatflt,cancel,tfrupee}
\usepackage{makecell}
\usepackage{tikz}
\usepackage{setspace}
\usepackage{color}
\usepackage{multirow}
\usepackage{adjustbox}
\usepackage{bbding}
\usepackage{physics}
\usepackage{colortbl}
\usepackage{xcolor}
\usepackage{pifont}
\usepackage[nointegrals]{wasysym}
\renewcommand\IEEEkeywordsname{Keywords}

\makeatletter
\def\ps@IEEEtitlepagestyle{%
  \def\@oddfoot{\mycopyrightnotice}%
  \def\@evenfoot{}%
}
\def\mycopyrightnotice{%
  {\footnotesize 979-8-3503-3015-1/23/\$31.00  \copyright 2023 IEEE \hfill}% <--- Change here
  \gdef\mycopyrightnotice{}% just in case
}

\let\old@ps@IEEEtitlepagestyle\ps@IEEEtitlepagestyle
\def\confheader#1{%
    \def\ps@IEEEtitlepagestyle{%
        \old@ps@IEEEtitlepagestyle%
        \def\@oddhead{\strut\hfill#1\hfill\strut}%
        \def\@evenhead{\strut\hfill#1\hfill\strut}%
    }%
    \ps@headings%
}
\makeatother

\confheader{%
    \parbox{17cm}{\centering 13th International Conference on Computer and Knowledge Engineering (ICCKE 2023), November 1-2, 2023, Ferdowsi University of Mashhad}
}

\begin{document}

        \title{Distilling Knowledge from CNN-Transformer Models for Enhanced Human Action Recognition}

\author{\IEEEauthorblockN{Hamid Ahmadabadi,
Omid Nejati Manzari,
Ahmad Ayatollahi}\\
School of Electrical Engineering, Iran University of Science and Technology, Tehran, Iran}
% use for special paper notices
%\IEEEspecialpapernotice{(Invited Paper)}

% make the title area
\maketitle

\begin{abstract}

This paper presents a study on improving human action recognition through the utilization of knowledge distillation, and the combination of CNN and ViT models. The research aims to enhance the performance and efficiency of smaller student models by transferring knowledge from larger teacher models. The proposed method employs a Transformer vision network as the student model, while a convolutional network serves as the teacher model. The teacher model extracts local image features, whereas the student model focuses on global features using an attention mechanism. The Vision Transformer (ViT) architecture is introduced as a robust framework for capturing global dependencies in images. Additionally, advanced variants of ViT, namely PVT, Convit, MVIT, Swin Transformer, and Twins, are discussed, highlighting their contributions to computer vision tasks. The ConvNeXt model is introduced as a teacher model, known for its efficiency and effectiveness in computer vision. The paper presents performance results for human action recognition on the Stanford 40 dataset, comparing the accuracy and mAP of student models trained with and without knowledge distillation. The findings illustrate that the suggested approach significantly improves the accuracy and mAP when compared to training networks under regular settings. These findings emphasize the potential of combining local and global features in action recognition tasks.
\end{abstract}

% no keywords

\section{Introduction}

Human action recognition is a dynamic study field in computer vision realm that encompasses both video-based and still image-based approaches. Video-based methods \cite{pareek2021survey} utilize temporal information to understand and classify human actions, while still image-based methods \cite{eshraghi2023cov} face the challenge of lacking such temporal cues. To counterbalance this limitation, action recognition of still images relies on alternative cues, including human pose \cite{liu2021normalized}, human-object interactions \cite{khaire2022deep}, salient regions \cite{al2019temporal}, backgrounds, and objects. These cues play a crucial role in various applications including abnormal behavior recognition, image search, and image labeling. Despite the absence of temporal information in still images, advancements in precise human pose estimation and effective interaction models have contributed significantly to the progress action recognition of still images. Conversely, video-based action recognition heavily relies on motion information, particularly optical flow features, which have proven to be discriminative and complementary to spatial cues. Understanding the strengths and limitations of these approaches is vital for researchers aiming to enhance the recognition and understanding of human actions across diverse practical applications. Detecting human actions from images has numerous applications in industries such as self-driving cars, security and surveillance systems, and human-robot collaboration, among others. These applications are now being realized using deep learning techniques.

The remarkable achievement of deep learning approaches in various applications has motivated researchers to extensively utilize these approaches in their investigations \cite{Dilated, LNL}. In the domain of computer vision, CNN and vision transformer networks(VIT) have emerged as the most prevalent deep networks, employing diverse architectures. One prominent technique employed in the training process of deep neural networks is Knowledge Distillation, which aims to enhance the performance and efficiency of smaller networks. Knowledge Distillation involves transferring knowledge from larger "teacher" models to smaller "student" models. While teacher models with numerous parameters and layers require significant computational resources, student models with fewer parameters and layers offer faster data processing. However, student models often exhibit lower accuracy. By leveraging the expertise of teacher models, Knowledge Distillation accelerates and improves the training of student models, resulting in more efficient and compact models suitable for resource-constrained environments. In summary, Knowledge Distillation empowers simpler student models with distilled knowledge from complex teacher models, leading to improved performance and speed in training and inference across various applications \cite{hinton2015distilling}.

In our proposed method, we utilize a Transformer vision network as the student model, along with a convolutional network as the teacher model. The convolutional network extracts local image features, while the Transformer vision network focuses more on global features in the image using an attention mechanism. In this study, we employ the method of knowledge distillation to inject the local features captured by the teacher model into the global features extracted by the student model meanwhile the training phase, resulting in improved performance and accuracy in human action recognition. Throughout all stages, pre-trained networks on the ImageNet database are utilized to enhance learning and prevent overfitting. Additionally, the Stanford 40 dataset \cite{yao2011human} is used in this research, where our proposed approach leads to a remarkable increase in accuracy compared to training networks in the regular setting.

 \begin{figure*}[!t]
 \centering
  \includegraphics[width=.9\textwidth]{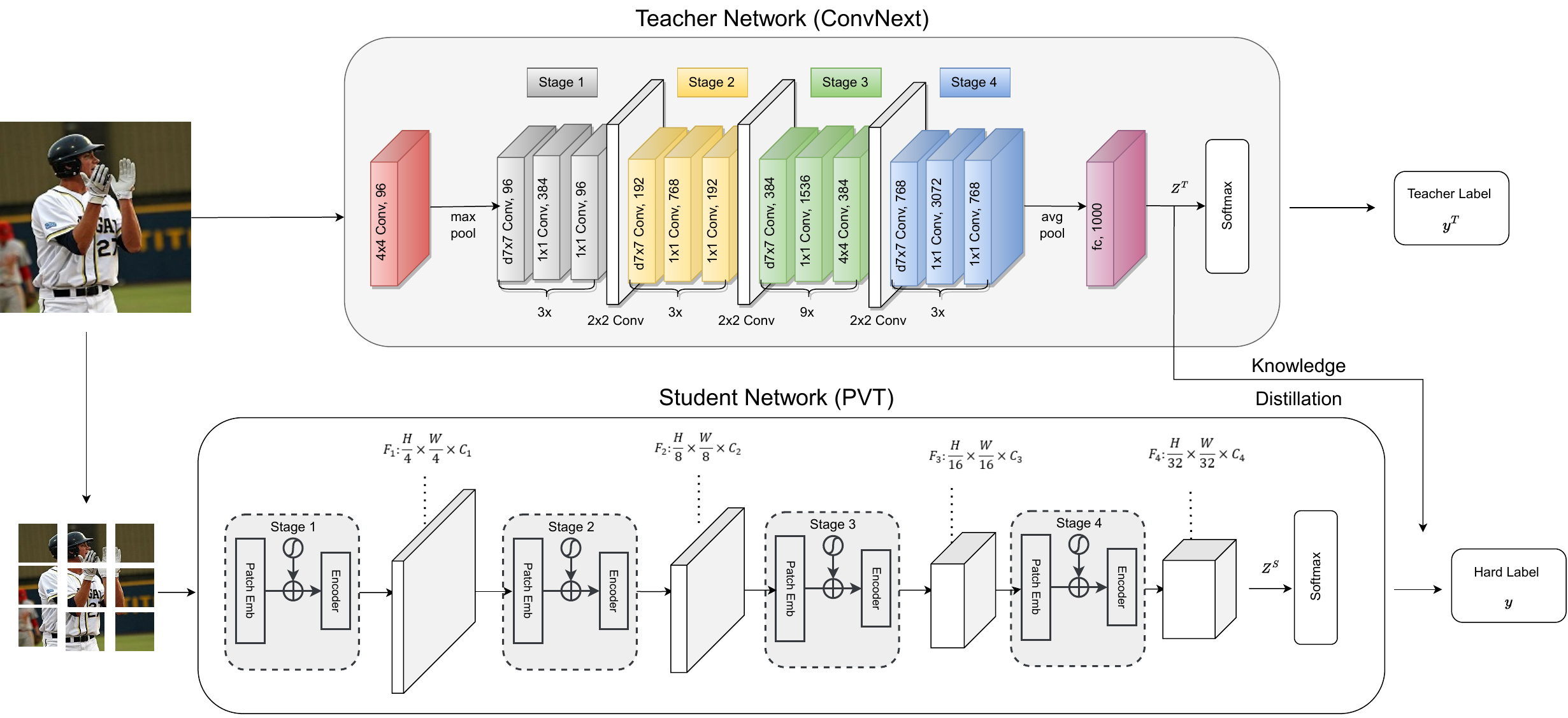}
  \caption{The general architecture of knowledge distillation. The structure of teacher model (ConvNeXt) is displayed at the top, while the architecture of student model (e.g. PVT) is shown at the bottom.}\label{fig.architecure}
  %\vspace{-4mm}
\end{figure*}

\section{METHODOLOGY}

In this section, we begin by introducing the knowledge distillation method, which serves as the core idea of our study, along with detailed explanations. Subsequently, we provide an overview of Vision Transformer (VIT) networks and their novel variants that have been employed as student models in this research. Finally, we delve into the introduction of ConvNeXt network and its advantages as a teacher model.

\subsection{Knowledge Distillation}

A deep learning technique called knowledge distillation is used to move knowledge from a larger, more complex model (the teacher) to a smaller, more effective model (the student). By utilizing the information and perceptions gathered by the instructor model, it seeks to enhance the performance of the student model (see Figure~\ref{fig.architecure}). Knowledge distillation's main principle is to teach the student model to behave in a way that is similar to that of the instructor model rather than immediately optimizing it for the job at hand \cite{xu2023teacher}. In the knowledge distillation process, several components and functions are involved:

\begin{itemize}
    \item \textbf{Teacher Model}: The model used by teachers to impart knowledge is often bigger and more precise. It has acquired intricate patterns and representations after being trained on a sizable dataset.
    \item \textbf{Student Model}: The learner model is a more compact and computationally effective model that tries to include the teacher model's information. By studying its predictions, it is taught to imitate the teaching model's actions.

    \item \textbf{Soft Targets}: Soft targets are the probabilities or logits produced by the teacher model for each input image. These soft targets provide rich information about the relationships between different classes and help the student model learn more effectively.

    \item \textbf{Hard Targets}: Hard targets refer to the ground truth labels or target values for the input images. These targets are used to help the student model learn by allowing it to compare its predictions with the actual labels while it is being trained.

\end{itemize}

The knowledge distillation process involves two main loss functions. The Knowledge Distillation Loss, denoted by the formula below, measures the discrepancy between the soft predictions of the learner model $\left(S_{i j}\right)$ and the soft targets provided by the expert model $\left(T_{i j}\right)$. It encourages the student model to produce similar output probabilities as the teacher model.

$$
L_{k d}=-\frac{1}{N} \sum_{i=1}^N \sum_{j=1}^C T_{i j} \log \left(S_{i j}\right)
$$

On the other hand, the Task-specific Loss, represented by the formula below, utilizes the hard targets $\left(y_i\right)$ and the predicted outputs of the student network $\left(\hat{y}_i\right)$ to guide the training phase. In this case, the loss function of cross-entropy is used.

\begin{equation}
L_{\text {task }}=-\sum_{i=1}^N y_i \log \left(\widehat{y}_i\right)
\end{equation}

The overall loss function used during training is a combination of the knowledge distillation loss and the task-specific loss, weighted by hyperparameters that control the trade-off between mimicking the teacher and optimizing for the task. To adjust the influence of the two loss functions, the total loss function $\left(L_{\text {total }}\right)$ combines them using a weighting factor $(\alpha)$ and a temperature parameter $\left(t^2\right)$. The formula is as follows:
\begin{equation}
L_{\text {total }}=(1-\alpha) \cdot t^2 \cdot L_{k d}+\alpha \cdot L_{\text {task }}
\end{equation}

\subsection{Overview of Vision Transformer Architecture}

The Vision Transformer (ViT) architecture has revolutionized computer vision by leveraging attention mechanisms to capture global dependencies. ViT partitions input images into fixed-size patches, which are then linearly projected using trainable linear projection layers. Positional context is incorporated into the structure by applying position embeddings to the transformed tokens, enabling the model to capture relative or absolute positional information.

To process the patch embeddings, ViT employs a Transformer encoder, composed of multihead self-attention (MHSA) and a feed-forward network (FFN) with layer normalization and residual connections. The MHSA applies attention mechanisms across different heads, allowing the model to learn various states of self-attention and effectively model relationships between patches. The outputs from several heads are combined and sent to the FFN for further processing. The following equations summarize the ViT operations:

\begin{equation}
\begin{aligned}
& x_0=\left[x_{\text {patch }} \| x_{\text {cls }}\right]+x_{\text {pos }} \\
& y_k=x_{k-1}+M H S A\left(\operatorname{LN}\left(x_{k-1}\right)\right) \\
& x_k=y_k+M L P\left(\operatorname{LN}\left(y_k\right)\right)
\end{aligned}
\end{equation}

However, due to the quadratic escalation in computational complexity with a rising number of tokens, vanilla ViT has constraints when performing jobs that demand high-resolution details. Advanced architectural designs have been suggested as a solution. For example, the Pyramid Vision Transformer (PVT) \cite{wang2022pvt} incorporates a hierarchical pyramid structure to effectively capture multi-scale information by combining lower-resolution features with their higher-resolution counterparts. Similarly, the Convolutional Vision Transformer (Convit) \cite{wu2021cvt} seamlessly integrates convolutional layers with transformer blocks, leveraging local feature extraction and global context modeling. The Multiscale Vision Transformer (MViT) \cite{li2022mvitv2} is a deep learning architecture specifically designed for computer vision tasks. It integrates the strengths of Vision Transformers (ViT) with the ability to extract multiscale features. The primary objective of
MViT is to effectively capture both local and global contextual information in images by processing them at various scales. Furthermore, the Swin Transformer \cite{liu2022swin} partitions input images into non-overlapping patches and utilizes shifted windows and local self-attention procedure to capture high-level features with scalability and efficiency. Lastly, the Twins architecture \cite{chu2021twins} leverages dual branches comprising both convolutional and transformer pathways to effectively capture local spatial information alongside global context. These remarkable vision transformer architectures have surpassed expectations in various computer vision tasks, pushing the boundaries of visual recognition and comprehension.

The Vision Transformer (ViT) architecture, along with its advanced variants such as PVT, Convit, MVIT, Swin Transformer, and Twins, has significantly contributed to the field of computer vision. These architectures demonstrate the power of attention mechanisms and the combination of convolutional layers and transformers in capturing global dependencies, modeling multi-scale information, and efficiently handling various vision tasks. Further research and comparisons are needed to evaluate their performance in specific applications and explore techniques like Knowledge Distillation to merge local and global features, ultimately enhancing network capabilities.

\subsection{Unleashing the Power of ConvNeXt as a Teacher}

For our research, we employed ConvNeXt \cite{fan2023lacn} as a powerful teacher model. ConvNeXt has gained significant recognition for its efficiency and effectiveness in a wide range of computer vision tasks and represents a state-of-the-art convolutional neural network architecture. The specific variant of ConvNeXt utilized in our study was specifically designed to address scenarios with limited computational resources while ensuring high performance. This tailored version optimizes the model's architecture and parameterization to strike a balance between model size and accuracy. By leveraging the remarkable local feature extraction capabilities of ConvNeXt, we achieved exceptional results in our investigation. ConvNeXt, as a teacher model, can be represented mathematically as follows:

\begin{equation}
  ConvNeXt (x) = Conv (ReLU ( BN (WeightedSum ((x_i)))))
\end{equation}

Here, $x$ represents the input tensor, Conv is the convolutional layer, ReLU denotes for the enhanced linear unit activation function, BN refers to batch normalization, and WeightedSum calculates the weighted sum of the input tensor elements. By incorporating ConvNeXt as the teacher model, which serves as a knowledgeable guide, we employed the technique of knowledge distillation to transfer its rich knowledge to the student models (Convit, MVIT, PVT, Swin Transformer, and Twins). The objective was to enhance the performance of these student models by leveraging the learned insights and high-level representations from ConvNeXt.

\begin{table*}[t]
    \caption{Performance for Human Action Recognition on the Stanford40.}
    \begin{adjustbox}{width=.75\textwidth,center}

    \begin{tabular}{l|ccll}
    \toprule
         Network & Parameters (M) & GFLOPS (G) & Accuracy (\%) & mAP (\%) \\

        \midrule ConvNeXt (Teacher) & 49.48 & 8.6 & 89.08 & 90.01 \\
        \hline Swinv2 (Student) & \multirow{2}{*}{27.59} & \multirow{2}{*}{4.3} & 83.44 & 83.08 \\
               Swinv2 (KD) &  &  & 85.01(\textcolor{orange}{1.57$\uparrow$}) & 83.57(\textcolor{orange}{0.49$\uparrow$}) \\

        \hline Twins (Student) & \multirow{2}{*}{23.56} & \multirow{2}{*}{2.8} & 83.98 & 81.74  \\
               Twins (KD) &  &  &  84.54(\textcolor{orange}{0.56$\uparrow$}) & 83.02(\textcolor{orange}{1.28$\uparrow$}) \\
        \hline Mvitv2 (Student) & \multirow{2}{*}{23.43} & \multirow{2}{*}{4.2} & 85.17 & 83.91  \\
               Mvitv2 (KD) &  &  & 86.65(\textcolor{orange}{1.48$\uparrow$}) & 87.05(\textcolor{orange}{3.14$\uparrow$})  \\
        \hline Convit (Student) & \multirow{2}{*}{27.36} & \multirow{2}{*}{4.5} & 85.88 & 86.76 \\
               Convit (KD) &  &  & 86.60(\textcolor{orange}{0.72$\uparrow$}) & 90.32(\textcolor{orange}{3.56$\uparrow$})  \\
        \hline PVT-v2 (Student) & \multirow{2}{*}{22.06} & \multirow{2}{*}{2.9} & 87.05 & 86.82 \\
               PVT-v2 (KD) &  &  & 88.59(\textcolor{orange}{1.54$\uparrow$}) & 90.49(\textcolor{orange}{3.67$\uparrow$}) \\
        \bottomrule
        \end{tabular}
    \end{adjustbox}
    %\vspace{-5mm}
\end{table*}

\section{EXPERIMENTAL RESULTS}

\subsection{Implementation Details}

The suggested solution was tested in the Google Colab environment and put into practice with the PyTorch framework. On the Stanford 40 dataset, which relates to the subtask of human action recognition, the effectiveness and efficiency of our suggested strategy were assessed. This dataset comprises of 9,532 images, which 4,000 images allocated for training and 5,532 images for testing. The input image size for both testing and training phases was resized to $224 \times 224$ pixels. Mean Average Precision (mAP) was measured for performance comparison and evaluation. To augment the training data, random cropping, resizing to 224×224 pixels, and horizontal flipping were applied. Initially, we trained the student and teacher networks separately. The cross-entropy loss function was utilized for training the models in normal conditions. The SGD optimizer was employed for loss function optimization, with learning rate, momentum, and weight decay values set to 0.001, 0.9, and 0.00001, respectively. A learning rate scheduler was used to lower the learning rate by a factor of $0.1$ every $10$ epochs while the teacher network was trained for $50$ epochs. Both the student networks, in normal and distilled modes, were trained for $15$ epochs. Similar to this, a scheduler for learning rates was used, which reduced learning rates by a factor of $0.3$ every five epochs. When employing the distillation method in the overall loss function, the $T$ and $alpha$ values were set to $0.45$ and $11$, respectively. It is worth noting that all networks utilized pre-trained weights obtained from the ImageNet database.

 \begin{figure}[!t]
 \centering
  \includegraphics[width=.9\columnwidth]{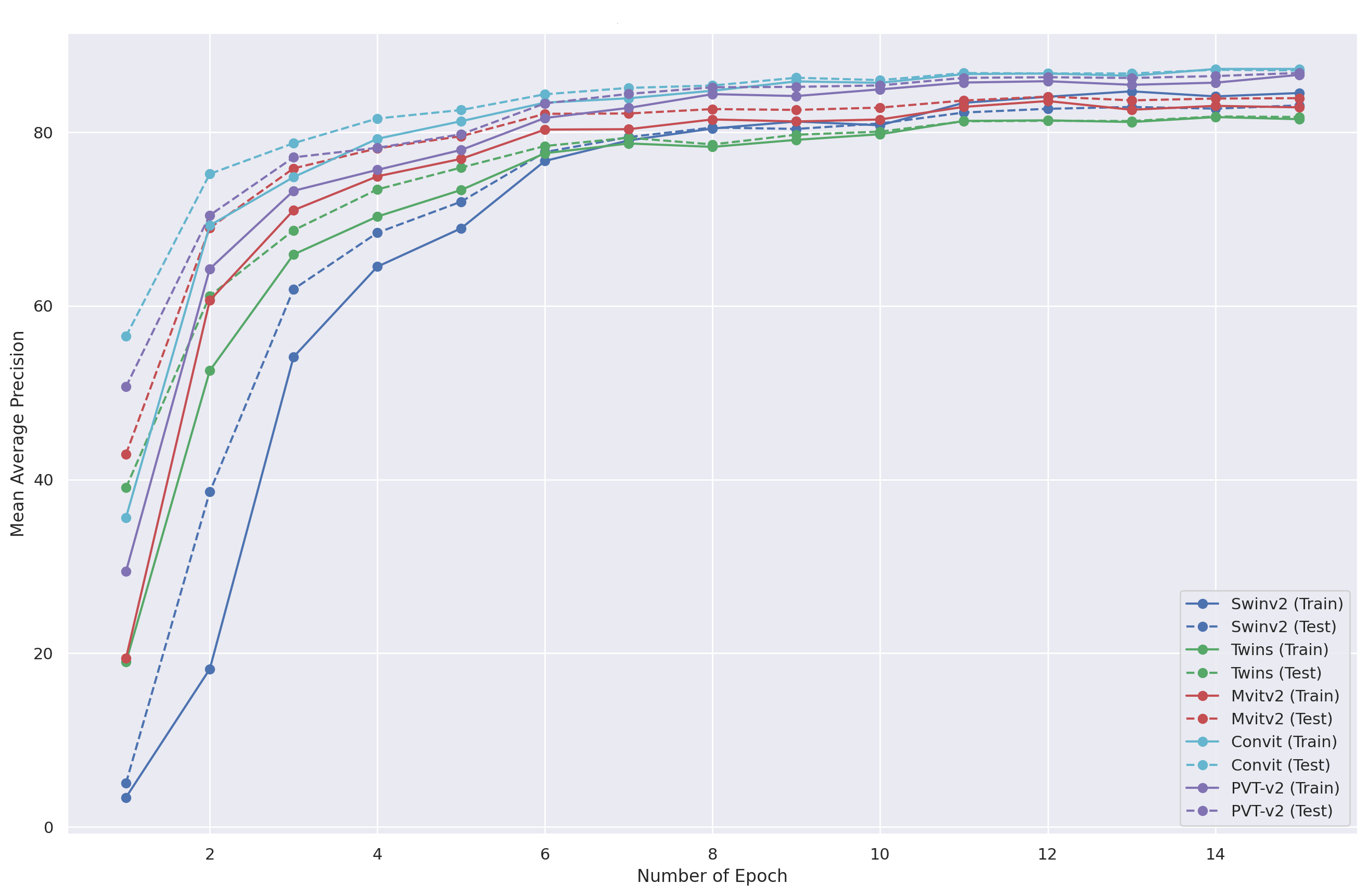}
  \caption{Accuracy of training and testing for five different models.}\label{fig.res1}
  %\vspace{-4mm}
\end{figure}

\subsection{Results and comparisons}

In this section, we demonstrate how distilling knowledge from a CNN to a Transformer can enhance the performance of Transformers by incorporating an inductive bias that leads to more CNN-like predictions.

In our study, we examined the effectiveness of our proposed method by training five different types of Transformer networks in two modes: regular training and training with a teacher using the knowledge distillation technique. Table 1 presents the features, results, and accuracies obtained from various Vision Transformer networks during both regular and teacher-student training. Furthermore, based on Figure 2, which illustrates the mAP with respect to the number of epochs in the regular training mode, it can be discerned that the PVT network exhibits superior performance and accuracy compared to other networks. Overall, Table 1 illustrates the impact of distillation on each of the five models we studied in the Human Action Recognition setup. Notably, the most significant improvement in both accuracy and mAP is observed when the student is a pure Transformer-based model. As the inherent recurrent biases of the basic-model student weaken, the extent of performance improvement increases. For example, PVT gains an mAP of 90.49\% and an accuracy of 88.59\%, surpassing other hybrid-Transformers such as Convit and Twins. Additionally, introducing the inductive bias results in a 1.57\% gain over the Swin model, representing the largest accuracy improvement among all networks.
It's important to point out that the teacher network, ConvNext, achieved 90\% accuracy on the test data.

In summary, Table 1 demonstrates that knowledge distillation, regardless of the setting, improves model performance. Moreover, for better visualization and comprehension of this matter, we have plotted mAP in Figure 3 for the best model, which is the PVT, in both regular training and Knowledge Distillation modes, for both test and training datasets. This graph distinctly illustrates the performance improvement of the proposed approach. 
To assess the extent to which the representations learned by our Hybrid-Transformer closely resemble those of the CNNs, we conducted a series of quantitative evaluations. We employed similarity scores, such as cosine similarity and Euclidean distance, to measure the similarity between feature vectors extracted from both models. Additionally, we performed feature-matching evaluations, comparing the activations of specific layers in the Hybrid-Transformer and CNNs. Our results consistently demonstrated high similarity scores and close alignment between the representations learned by the Hybrid-Transformer and the CNNs, validating the claim that distillation from a CNN to a Hybrid-Transformer produces representations that closely resemble the teacher model. This stark contrasted with experiments using CNNs as teachers and pure Transformers as students, where we observed significant differences in representation quality.

 \begin{figure}[!t]
 \centering
  \includegraphics[width=.9\columnwidth]{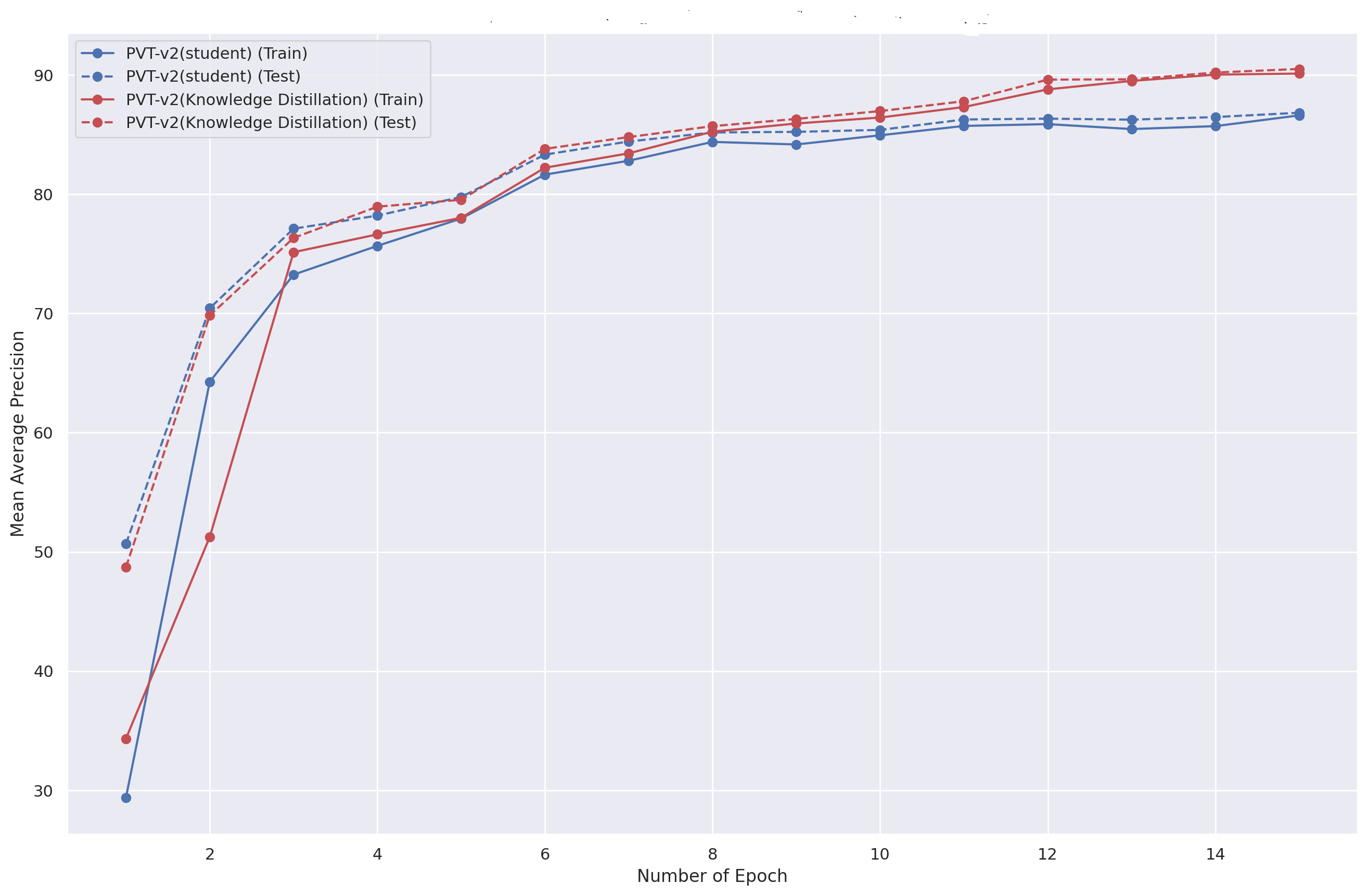}
  \caption{Accuracy of training and testing for PVT model.}\label{fig.res1}
  %\vspace{-4mm}
\end{figure}

\section{Conclusion}

This research paper introduced a study that aimed to enhance human action recognition by integrating knowledge distillation with CNN and ViT models. The proposed approach utilized a Transformer vision network as the student model and a convolutional network as the teacher model. To improve the performance and efficiency of the student models, the information from the expert model was transmitted to them. Experimental results on the Stanford 40 dataset showcased a significant improvement in accuracy and mean average precision compared to traditional training methods. These findings emphasize the potential of blending global and local features for action recognition tasks. In future investigations, it would be beneficial to explore the application of these techniques in other computer vision tasks and datasets.

\bibliographystyle{IEEEtran}

{\small
\bibliography{article}}

\end{document}